# A recursive self-improvement framework for electron microscopy an alysis

**Authors:** Yaotian Yang[1]†, Sijia Tao[2]†, Yuping Zhang[2], Jiangjie Qiu[1], Qinghua Zhang[2]*, Xiao Chen[1,3]*, Fei Wei[1,3]

**Affiliations:**

[1]Department of Chemical Engineering, Tsinghua University, Beijing 100084, China.

[2]Institute of Physics, Chinese Academy of Sciences, Beijing 100190, China.

[3]Ordos Laboratory, Ordos, Inner Mongolia, 017000, China.

*Corresponding author. Email: chenx123@mail.tsinghua.edu.cn (X.C.), zqh@iphy.ac.cn(Q.Z.)

†These authors contributed equally to this work.

## Abstract

Resolving local atomic structures such as defects and interfaces from electron microscopy contrast relies he avily on researcher judgement, leading to low efficiency and subjective standards. Here we present PhyEvo-Mic -RSI, a physics-guided and expert-informed recursive self-improvement framework that turns one-off human co rrections into reusable, verifiable capabilities by updating executable operators, parameters and routing rules wit h versioned inheritance. Given researcher-defined tasks, the system selects analysis paths from image features w ithout expert labels, while testing and inheriting code updates under fixed budgets. Combining physical priors w ith expert feedback improves atomic point-defect detection over larger models and raises cross-material interfac e coverage from 5.13% to 72.11% within a 10% area budget. Adapting the framework to micrographs from unse en studies further separates touching nanoparticles by aligning optimization with particle-level measurements. T his work moves computational microscopy beyond standalone models and static scripts towards self-evolving pl atforms that accumulate human scientific expertise.

Transmission and scanning transmission electron microscopy (TEM/STEM) enable direct observation of point defects, phase interfaces, nanoparticle morphologies and lattice distortions within materials[1-10]. Reliable physical measurements from microscopy images, however, require more than identifying anomalous pixels. Because local contrast can vary markedly—or even reverse—with composition, zone-axis orientation, thickness and signal-to-noise ratio, resolving aperiodic defects, heterogeneous interfaces and local lattice distortions relies heavily on subjective researcher judgement. This reliance leads to low throughput, inconsistent standards across operators, and isolated single-image corrections whose underlying physical criteria and parameter adjustments are lost once a task ends. Establishing objective, reusable analytical criteria is therefore critical for turning high-resolution imaging into reproducible, quantitative structure–property characterization.

Deep learning and training-free physical analysis offer complementary approaches[7, 11-19]. While deep neural networks capture complex texture mappings, sparse non-exhaustive expert annotations in microscopy often cause purely data-driven models to suppress unlabelled targets as background and degrade across materials[20]. Conversely, methods based on Fourier spectra, lattice symmetry or geometric phase provide interpretable structural references without training[21, 22], but their assumptions of periodicity and local regularity often fail to capture the boundaries of complex defects or atypical interfaces. Improving microscopy analysis therefore requires not only training recognition networks, but also coordinating physical and learned evidence according to image features, combining candidate geometries and determining when each operator is applicable[23, 24].

Scientific agents and automated algorithm discovery have recently opened new routes to reusable research workflows[25-28]. Multi-tool agents can orchestrate microscopy image-processing and materials-computation software[29-32]. Automated Design of Agentic Systems (ADAS)[33] and scientific operator-evolution frameworks go further by searching executable code to optimize data-processing pipelines[34-36], while hyperagents seek to make the improvement mechanism itself evolvable[37]. Applying framework-level RSI to materials microscopy nevertheless raises three fundamental questions. First, can sparse expert geometric feedback and physical evidence produce algorithmic gains beyond a frozen reference system, rather than increasing recall simply by relaxing the selection budget? Second, can later iterations inherit new methods while retaining existing capabilities, instead of replacing the parent system with a single high-scoring result that degrades earlier scenarios? Third, can controlled ablations and cross-task tests distinguish redundant code from effective mechanisms and align optimization metrics with downstream physical measurements?

To address these challenges, we developed PhyEvo-Mic-RSI, a physics-guided and expert-informed framework that makes microscopy analysis methods themselves the target of recursive self-improvement. By integrating physical diagnostics, expert feedback and autonomous program evolution, the framework enables analysis capabilities to be continually refined, validated and inherited across iterations, rather than remaining fixed or confined to individual tasks. We demonstrate cumulative capability improvement in atomic defect and interface analysis, followed by cross-task adaptation to nanoparticle instance separation, where the framework accommodates distinct imaging characteristics and measurement objectives.

# Results

### 1. An executable and inheritable architecture for microscopy analysis

PhyEvo-Mic-RSI uses image-analysis results and human corrections as the starting point for subsequent methodological improvement, preventing expertise from being lost when a one-shot analysis ends (Fig. 1a). In the analysis layer (Analyse; Fig. 1b), raw images and optional crystallographic priors (such as projected CIF templates) are processed in parallel by training-free physical operators—extracting periodicity, spectral, phase-residual and symmetry cues—and a supervised student detector to generate candidate geometries for expert review.

When existing methods perform poorly under particular structures or contrast conditions, the research layer (Research; Fig. 1b) initiates a self-improvement cycle using development-side observations. The diagnostic module reads the current system state and signals such as structural features, reliability distributions and disagreement between physical and learned branches. It identifies failure modes and proposes executable changes to analysis operators, student training or routing rules. Every candidate program is executed independently in an isolated environment and evaluated for improvement and capability retention under the same candidate budget. The state layer (Remember; Fig. 1b) enables capabilities to accumulate across iterations. At iteration $t$, the system state is $S_t = \{B_t, W_t, A_t, E_t\}$, where $B_t$ is the composable analysis-capability library, $W_t$ contains student-network parameters or numerically fitted state, $A_t$ specifies the inference-time rules for deciding when to invoke each capability, and $E_t$ records proposal code, evaluation results and parent–child lineage. While routine inference reads only active $\{B_t, W_t, A_t\}$, each research evaluation updates the parent state upon admission, appends all trial outcomes to Et, and retrieves both to guide subsequent diagnostics.

Routine inference is decoupled from method evaluation through separate adoption (Fig. 1c) and admission (Fig. 1c) rules. At inference, $A_t$ routes analysis paths using only observable structure, reliability and branch-disagreement signals without expert labels. During evaluation under a fixed 10% image-area budget, admission to $S_{t+1}$ requires ≥+ 0.2 pp（percentage-point）gains in both three-source equal-weight coverage $J$ and non-fitted source mean $\Delta J_{\text{non-fit}}$, alongside source-retention, rotation/reflection, rotation-consistency and reverse-order replay checks.

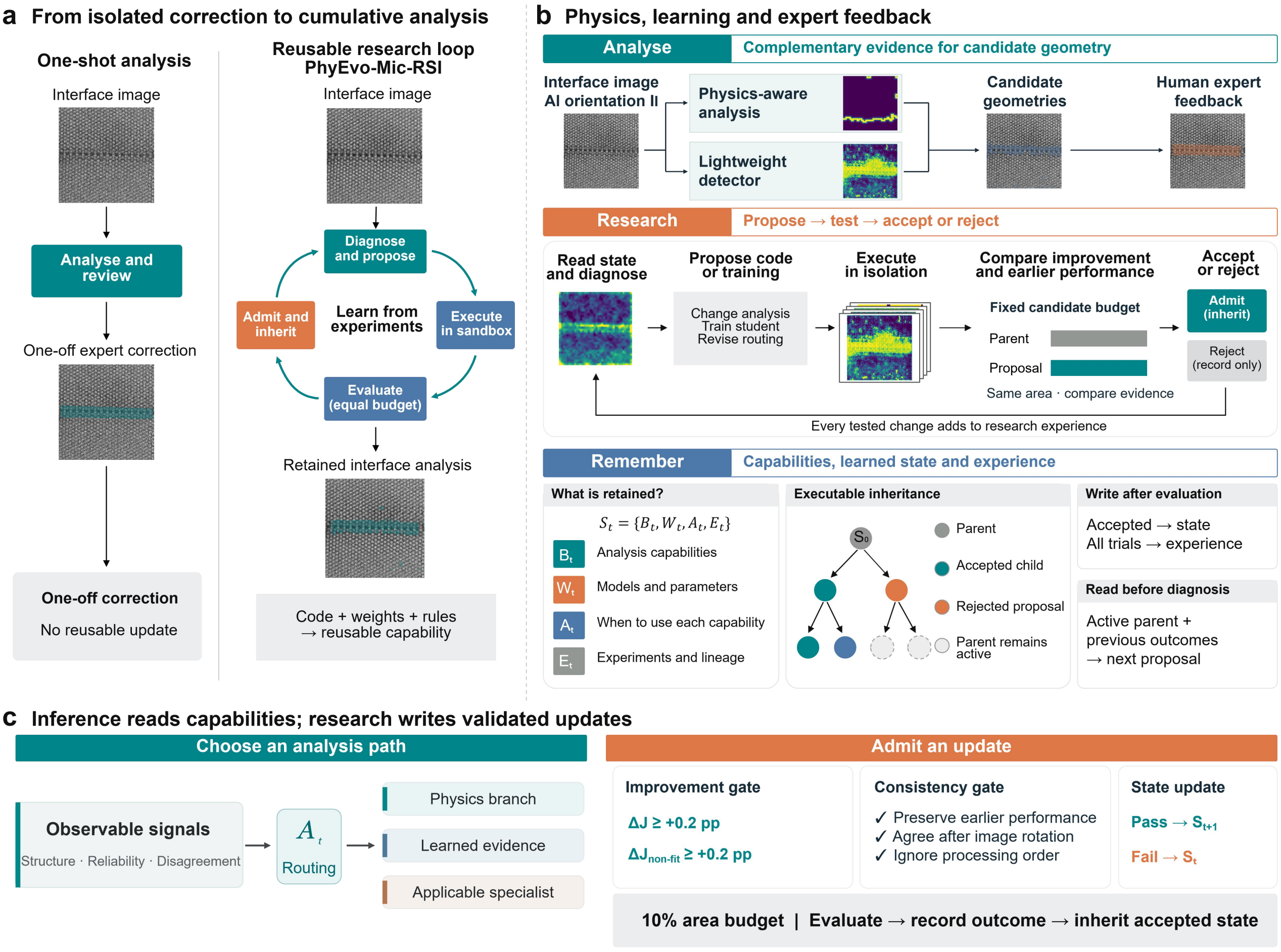

**Fig. 1 | System architecture and state updates in PhyEvo-Mic-RSI. a,** One-shot image correction versus a reusable research cycle that accumulates validated code, learned parameters and adoption rules. **b,** The Analyse, Research and Remember layers. Physical and learned branches generate candidate geometries for expert review; research diagnoses failures, proposes executable changes, tests them in isolation and compares them with the active parent. The retained state $S_t = \{B_t, W_t, A_t, E_t\}$ stores analysis capabilitie

s, learned parameters, adoption rules and experimental records and lineage. Accepted proposals update the active capabilities; all trial outcomes contribute to research experience. **c,** Inference-time path adoption (left) and research-time update admission (right). Routing uses observable structure, reliability and branch disagreement without expert labels. For the interface-analysis example, admission requires gains of at least 0.2 percentage points (pp) in both equal-source coverage and the mean over fit-excluded sources under a fixed 10% image-area budget, together with source-retention, rotation-consistency and order-independent replay checks. Pass/fail denotes updating or retaining the active parent; trial outcomes are recorded in either case.

## 2. Building foundational capabilities from physical priors and expert corrections

Because training-free physical operators rely on idealized periodicity and contrast assumptions, learning solely from their outputs propagates missed targets to the student model. We therefore introduced direct expert review of raw images as an independent supervision signal (Fig. 2a, full composition is given in Methods section 3). Across 154 reviewed microscopy cases, two domain experts corrected geometries in 152 cases—100 point-defect cases containing 808 labelled sites, 51 interface cases and one disordered-region case—and identified positive defects in 21 cases where physical-teacher predictions were empty, demonstrating that human feedback must recover missed targets rather than merely filter existing candidates.

We combined these independent human corrections with outputs from training-free physical operators (referred to as the physical teacher) to train a lightweight convolutional encoder–decoder (the student network; Fig. 2a). Human positive annotations, physical-teacher positive pseudo-labels and the weaker teacher-background prior were treated separately. Unlabelled pixels remained unknown rather than becoming human-labelled negatives. An exclusion halo around each newly annotated positive point masked conflicting weak teacher-background supervision, preventing sparse new positives from being suppressed during training (Fig. 2a). Trained on the same 151 non-$WSe_2$ cases (Methods), this compact student achieved 64.31% Point F1 on 107 unseen $WSe_2$ images, far exceeding the same architecture trained with human labels alone (3.73%) as well as much larger retrained U-Net[38] (27.46%) and EMSeek Ref-UNet (35.89%) baselines. Reference points, matched predictions and unmatched predictions in a shared test field illustrate the differences in missed and spurious detections (Fig. 2b, bottom).

To translate learned detection into executable adoption rules without retraining student weights, we partitioned the 107 $WSe_2$ images into 32 fitting, 22 selection and 53 confirmation images (Methods section 4). On the 53 confirmation images (Fig. 2b, top), the RSI-retained program achieved 78.71% Point F1, outperforming the original physical branch (71.95%), frozen student (64.65%), fixed 50:50 quota fusion (68.82%) and same-label physics-only calibration (74.82%). Student-based methods were averaged across three seeds, whereas the two physical controls were deterministic. It also exceeded the calibration control, fitted using the same labels but without variable student evidence, by 3.89 pp. The improvement arose from using student responses and local image features to rerank physical candidates, replace a controlled fraction and remove low-confidence tail candidates. Program-level evolution of adoption rules can therefore make better use of existing model outputs without retraining the network.

Within the overall capability-evolution process of PhyEvo-Mic-RSI, physical analysis, student learning and geometric routing progressively contributed complementary gains across five development sources (Fig. 2c). On non-periodic SuperScan and $MoWTe_2$ point tasks where physical baselines failed (0.00%), the student network recovered 28.82% and 14.29% positive coverage, respectively (Fig. 2c, left). Across the three interface sources ($ZrO_2$, Al orientation I and Al orientation II) under a fixed 10% area budget, equal-source coverage rose from 5.13% for initial physical methods (12.36%, 2.77% and 0.26%) to 26.09% and 27.10% after introducing the student and early fusion (raising Al II to 47.62% while Al I remained at 4.00%). Geometry-aware routing then boosted $ZrO_2$ to 74.44% and Al I to 58.34% (three-source mean, 59.76%), culminating in an 18-branch frozen refe rence system at 67.95% (76.49%, 63.17% and 64.19%; Fig. 2c, right) that anchored the final stage of automated framework-level evolution.

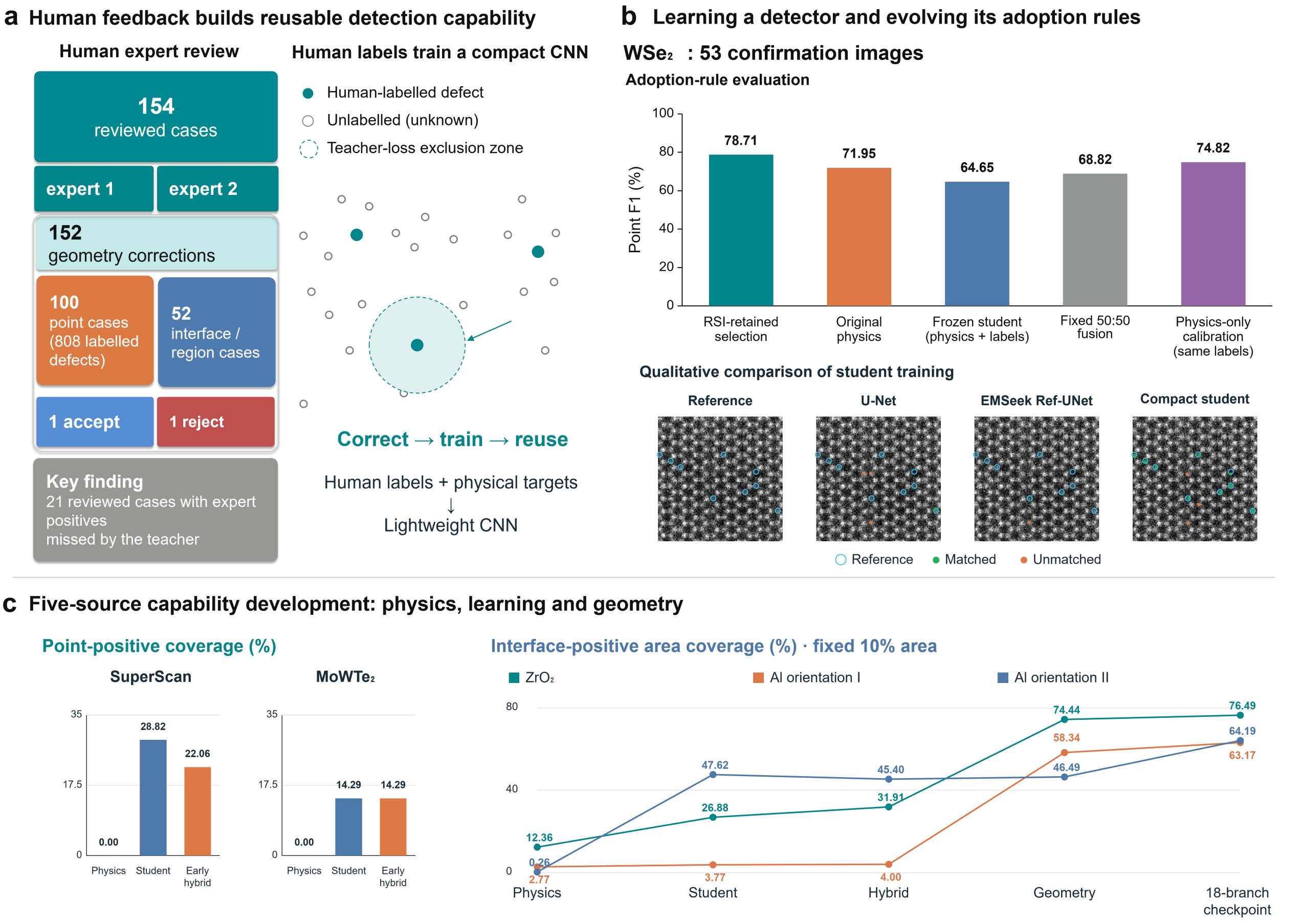

**Fig. 2 | Physical priors, expert supervision and geometric routing establish foundational analysis capabilities.** a, Human expert review (154 reviewed cases and 152 geometry corrections, including 21 cases with human positives despite empty physical predictions) and the teacher-loss exclusion zone, which masks conflicting weak teacher-background signals around new positives while leaving unlabelled regions unknown. b, Top: point F1 on 53 $WSe_2$ confirmation images (7,969 reference points), comparing RSI-retained selection (78.71%), original physics (71.95%), the frozen student trained with physics and human labels (64.65%), fixed 50:50 fusion (68.82%) and same-label physics-only calibration (74.82%). Bottom: qualitative comparison in the same test field (seed 23) showing reference points (cyan open circles), matched predictions (green) and unmatched detections (orange) for retrained U-Net, EMSeek Ref-UNet and the compact student. c, Five-source capability development. Left: positive-point coverage for SuperScan and $MoWTe_2$. Right: interface coverage for $ZrO_2$, Al orientation I and Al orientation II under a fixed 10% area budget, progressing from the physical baseline (three-source mean, 5.13%) through student learning (26.09%), early hybrid fusion (27.10%) and geometric routing (59.76%) to the 18-branch checkpoint (67.95%).

## 3. Multi-trajectory program improvement and state inheritance for interface analysis

Across successive stages of capability evolution—incorporating physical analysis, expert-informed student learning, geometric routing and executable program updates—PhyEvo-Mic-RSI increased equal-source interface coverage from 5.13% to 72.11% under a fixed 10% image-area budget (Fig. 3a, top). The initial capability-development stages increased coverage from 5.13% to 67.95%, followed by a separately audited autonomous program-evolution stage that further improved coverage to 72.11% through recursive program updates and state inheritance. Starting from the shared 18-branch intermediate state ($J = 67.95\%$), we launched three independent trajectories: a domain–residual route with 10 proposals, a learned-context route with 12 proposals and a boundary-shell route with 14 proposals. Their source-specific trajectories reveal complementary improvements across materials (Fig. 3a, bottom). Three admitted updates in the domain–residual route, at rounds 5, 8 and 9, raised $ZrO_2$ interface coverage from 76.49% to 88.07%, a source-specific gain of 11.58 pp. Round 8 contributed a cross-sourc e gain of 0.66 pp on Al I, and the best retained system from this route reached $J = 72.11\%$, an overall gain of 4.16 pp. The learned-context route admitted an update at round 11, raising Al II from 64.19% to 65.78% (+1.59 pp; overall $J = 68.62\%$, +0.67 pp). Four admitted updates in the boundary-shell route, at rounds 5, 11, 12 and 13, advanced Al II from 64.19% to 67.66% (+3.47 pp; overall $J = 70.46\%$, +2.51 pp); round 5 contributed 0.72 pp on Al I.

These improvements were not isolated corrections. Lineage records for all 36 main-trajectory proposals, together with 3 single-step routing references (39 proposals in total; 8 accepted and 31 rejected), show that later programs reused earlier admitted changes through parent-program calls and numerical-state reads (Fig. 3b). In the domain–residual route, round 5 introduced grain-boundary domain-transition-band reranking over the frozen parent. After rounds 6 and 7 were rejected, the system retained the round-5 state. Round 8 added a gated branch-disagreement residual operator, and round 9 used its output for orthogonal inverse-residual completion; the resulting terminal state was subsequently invoked by round 10. Similarly, coherent ridge exchange at round 5 of the boundary-shell route provided the starting point for shell expansion, anchor continuation and frontier extension at rounds 11, 12 and 13. The code and numerical state of each admitted node participated in subsequent inference, making earlier improvements part of later analysis.

Under a fixed exchange quota of 1,259 replaceable pixels per transition, the three successive boundary-shell updates (rounds 5→11, 11→12 and 12→13) consistently recovered more expert-positive pixels than they displaced, yielding consecutive net gains of 493, 404 and 319 positive pixels (Fig. 3c).

To assess whether the improvements extended beyond representative cases, we evaluated all 8 admitted nodes on 51 image planes, giving 408 program–plane combinations and 2,040 comparison records (Fig. 3d). Relative to a proximity-ranking control matched for support, total selected-pixel count and exchange quota, all 8 admitted nodes achieved positive equal-source net gains, ranging from 0.22 to 2.20 pp. Among the 408 plane-level comparisons, 123 improved, 277 were unchanged and 8 declined. The largest single-plane decreases were 0.26 pp on Al I in the domain route and 5.44 pp on Al II in the boundary-shell route. The archived systems also passed 918 horizontal-reflection and 180°-rotation prediction-consistency checks. Across 153 isolated replays, recomputed scores and support masks exactly matched the archived outputs, demonstrating reproducibility for the same inputs and consistent responses to image transformations.

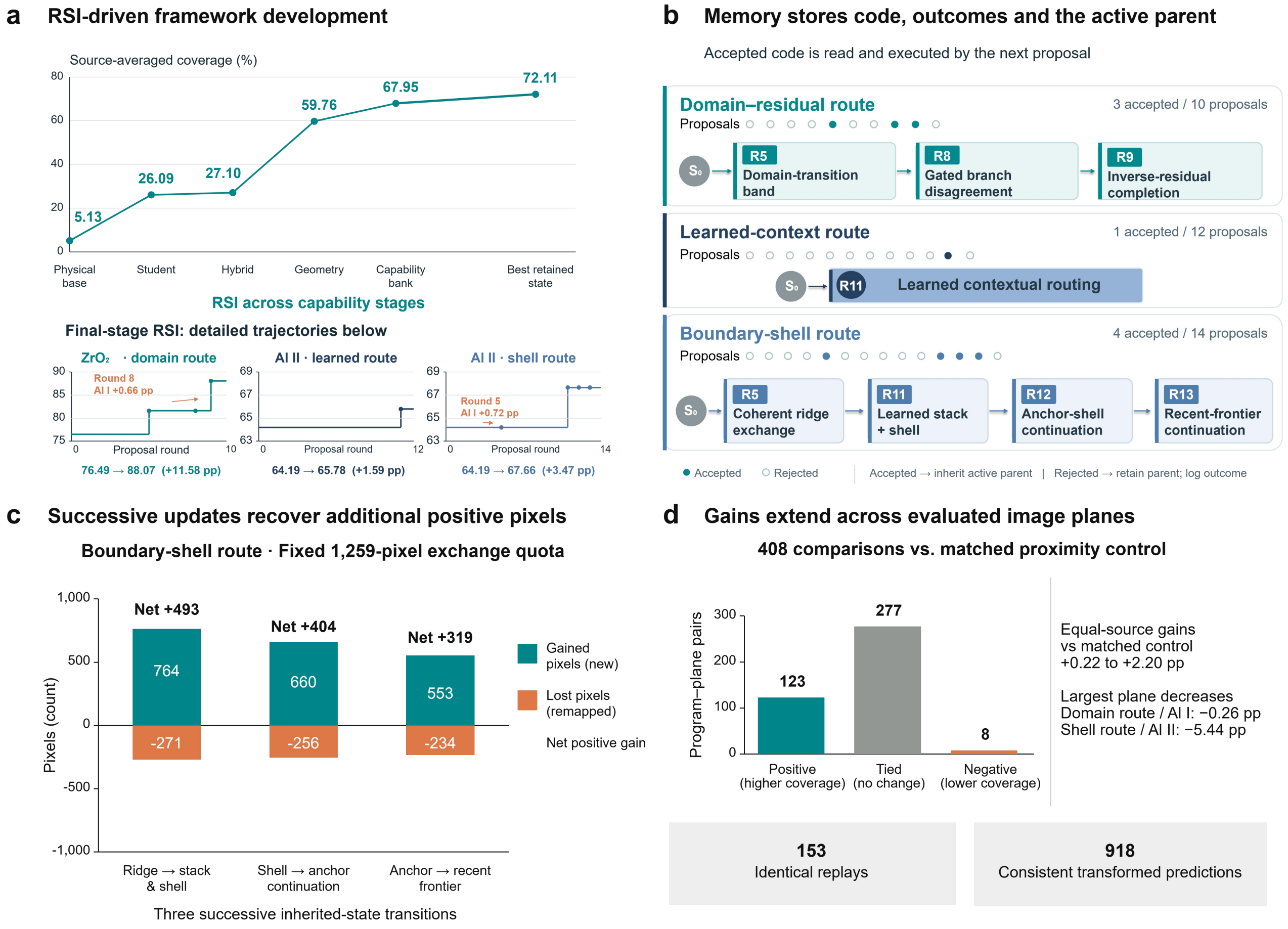

**Fig. 3 | Capability evolution and cross-iteration state inheritance in the RSI framework.** a, Top: source-averaged coverage across the three interface sources progresses through the physical base (5.13%), student learning (26.09%), hybrid fusion (27.10%), geometric routing (59.76%), the capability bank (67.95%) and the best retained state (72.11%). Bottom: final-stage retained-state step curves for the focal sources of three independent trajectories ($ZrO_2$ improves from 76.49% to 88.07%; Al II improves from 64.19% to 65.78% and 67.66% along two routes; annotations indicate cross-source gains on Al I at rounds 8 and 5). b, Proposal lineage and sequential state inheritance across the three routes. Open circles denote rejected proposals (retaining the active parent and logging the outcome); filled circles and connected cards (R5–R13) denote accepted updates that inherit and update the active parent state from $S_0$. c, Three successive boundary-shell transitions yield net positive gains of 493, 404 and 319 expert-positive pixels under a fixed exchange quota of 1,259 pixels per transition. d, All 8 admitted nodes across 51 planes (408 program–plane pairs) compared with the matched proximity control: 123 improve, 277 tie and 8 decline, alongside 153 identical isolated replays and 918 consistent transformed predictions.

## 4. Mechanistic attribution of gains and comparison of search strategies

Under the fixed 10% candidate-area budget, RSI redirected quota from non-interface regions towards grain-boundary transition bands. Coverage on $ZrO_2$ increased from 76.49% to 88.07%, recovering 4,446 additional positive pixels without losing any previously covered positives, while coverage on a representative Al II plane rose from 63.52% to 68.15% (Fig. 4a). Mechanistic interventions showed that these gains depended on specific physical computations and accumulated state. Removing domain correction reduced coverage by 5.05 pp. Under matched exchange quotas, removing inverse-residual evidence and resetting the parent to the initial system reduced coverage by 1.32 and 3.49 pp, respectively. By contrast, removing component gating or cross-scale aggregation had almost no effect on the shell update's net gain (Fig. 4b). These controls distinguish effective mechanisms from redundant details. Source-specific gains across all 8 admitted updates further demonstrate complementary contributions of different operators to the three interface sources (Fig. 4c).

To examine the role of research organization, we compared three strategies over 108 proposal opportunitie s using the same starting point, $J = 67.95\%$, and identical admission rules (Fig. 4d). Generic iterative coding an d replacement-based updates achieved terminal coverage of 71.17% and 71.30%, respectively, exceeding the 68. 60% achieved by structured inheritance. Across 36 opportunities per group, these two less-constrained strategies admitted 12 and 11 updates versus 5 for structured inheritance, while all 12 execution failures were safely isolat ed without altering parent states.

Beyond comparing predefined search strategies, we examined whether the research strategy itself could ev olve. In a separate paired experiment, a revised history-informed strategy achieved 38.1081% coverage from a s hared starting point of 35.0053%, compared with 35.7218% under the original strategy. Across two paired starti ng states, the revised strategy showed a mean advantage of 1.1931 percentage points under the predefined criteri on. The revised strategy was subsequently adopted, although its next proposed program failed the independent p rogram-admission threshold. This demonstrates that research strategies can be updated and inherited independen tly of individual analysis programs.

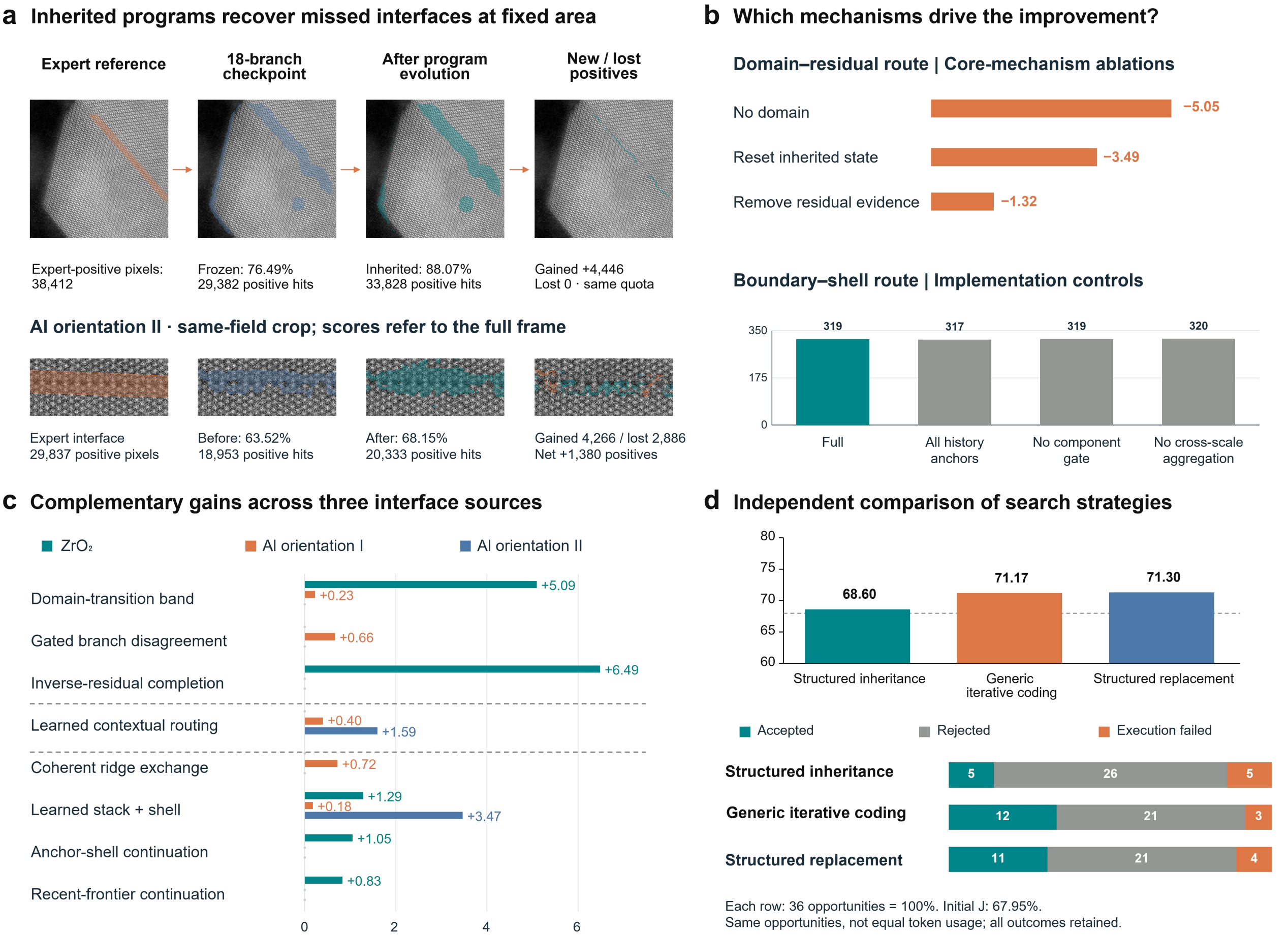

**Fig. 4 | Spatial attribution, mechanistic ablations and equal-opportunity search-strategy comparisons for interface analysis.** a, Interface selections under a fixed 10% area quota. Top: full-field $ZrO_2$ coverage rises from 76.49% at the 18-branch checkpoint to 88.07% after program evolution, gaining 4,446 positive pixels and losing none. Bottom: same-field crops from Al orientation II; full-frame coverage rises from 63.52% to 68.15%, with a net gain of 1,380 positive pixels. b, Core-mechanism ablations and imple mentation controls. Top: in the domain–residual route on $ZrO_2$, removing domain correction, resetting the inherited state or remov ing residual evidence reduces coverage by 5.05, 3.49 and 1.32 pp, respectively. Bottom: in the boundary–shell route at a fixed exch ange quota of 1,259 pixels, net positive-pixel gains remain nearly unchanged across the full program (319), all historical anchors (3 17), removal of the component gate (319) and removal of cross-scale aggregation (320). c, Source-specific coverage gains (pp) of a ll 8 admitted updates relative to their respective parents, grouped by trajectory. d, Independent comparison of search strategies acro ss 108 matched proposal opportunities (3 groups × 36 opportunities; initial $J = 67.95\%$), showing terminal mean $J$ values (top: 68.

60% for structured inheritance, 71.17% for generic iterative coding and 71.30% for structured replacement) and accepted, rejected and failed proposal counts (bottom).

## 5. Cross-task adaptation to nanoparticle instance separation

To examine whether the framework could be repurposed beyond atomic-scale defect and interface localization, we adapted PhyEvo-Mic-RSI to nanoparticle instance separation in electron micrographs gathered from published studies (Fig. 5). This task changes both the image regime and the measurement objective: high foreground overlap is insufficient if touching particles are merged, since individual particle identities and counts must be preserved. We therefore evaluated foreground overlap alongside instance-level detection, boundary accuracy and counting error, with source papers separated across dataset splits [39, 40].

Using images from 18 source papers in the Electron Microscopy Particle Segmentation (EMPS) dataset [41], we assigned 21 images from 8 papers to fitting, 13 from 5 papers to development and 14 from 5 papers to an initial holdout (Fig. 5a; Methods section 8). Starting from Otsu thresholding[42], three four-proposal adaptation trajectories admitted 11 of 12 candidate updates under a foreground-IoU objective. Yet on the initial holdout, mean foreground intersection-over-union (IoU) increased only from 47.34% to 50.88%, instance F1 changed little (25.16% to 26.06%), and mean absolute particle-count error worsened from 10.65 to 30.68. Foreground expansion had merged adjacent particles, revealing that improving pixel-level segmentation did not necessarily improve particle-level quantification.

We therefore examined whether measurement-aligned selection could correct this mismatch. Twelve archived programs were compared with predefined conditional and marker-controlled watershed configuration libraries[43], with selection rules fixed before evaluation on a separate confirmation set of 25 images from 9 additional source papers (Fig. 5b; Methods section 8). Relative to Otsu, a conditional selector increased instance F1 from 25.13% to 34.18% and reduced relative count error from 0.89 to 0.74. Joint-endpoint watershed selection (WS / joint) achieved 75.77% foreground IoU and 42.24% foreground boundary F1, with an instance F1 of 53.66%, compared with 49.87% for IoU-only watershed selection (WS / IoU) and 25.13% for Otsu. Relative count error was reduced from 2.71 (WS / IoU) to 1.10 (WS / joint), although it remained above the Otsu value of 0.89. Thus, a large gain in particle-instance recognition was obtained by evaluating object-level measurement criteria rather than foreground overlap alone.

In a separate diagnostic case, qualitative expert feedback triggered a targeted, agent-driven refinement on the previously inspected G14 image of densely touching nanocubes (19 reference particles; Fig. 5c). Initial distance-based partitioning produced only 11 predicted instances (instance F1, 33.33%; internal-seam boundary F1, 54.17%). After feedback that the merged boundaries were implausible, the research agent identified the neglected dark seams between particles and introduced image-derived centre markers and intensity-guided watershed topography. Instance F1 rose to 87.80% (18 one-to-one matches among 22 predictions), and internal-seam boundary F1 reached 72.01%. Separating coarse centre detection from finer seam tracing further increased internal-seam boundary F1 to 85.29%, while instance F1 remained 87.80% and foreground IoU remained 95.15%. This single-image intervention demonstrates a feedback-driven change in the analysis mechanism and was not part of the independent-source confirmation.

Together, the independent-source confirmation and the diagnostic refinement illustrate how the same propose–execute–evaluate–inherit framework can be redirected to a distinct electron-microscopy analysis task. In this setting, productive improvement required moving beyond foreground coverage toward the physically relevant requirement of separating and counting individual particles.

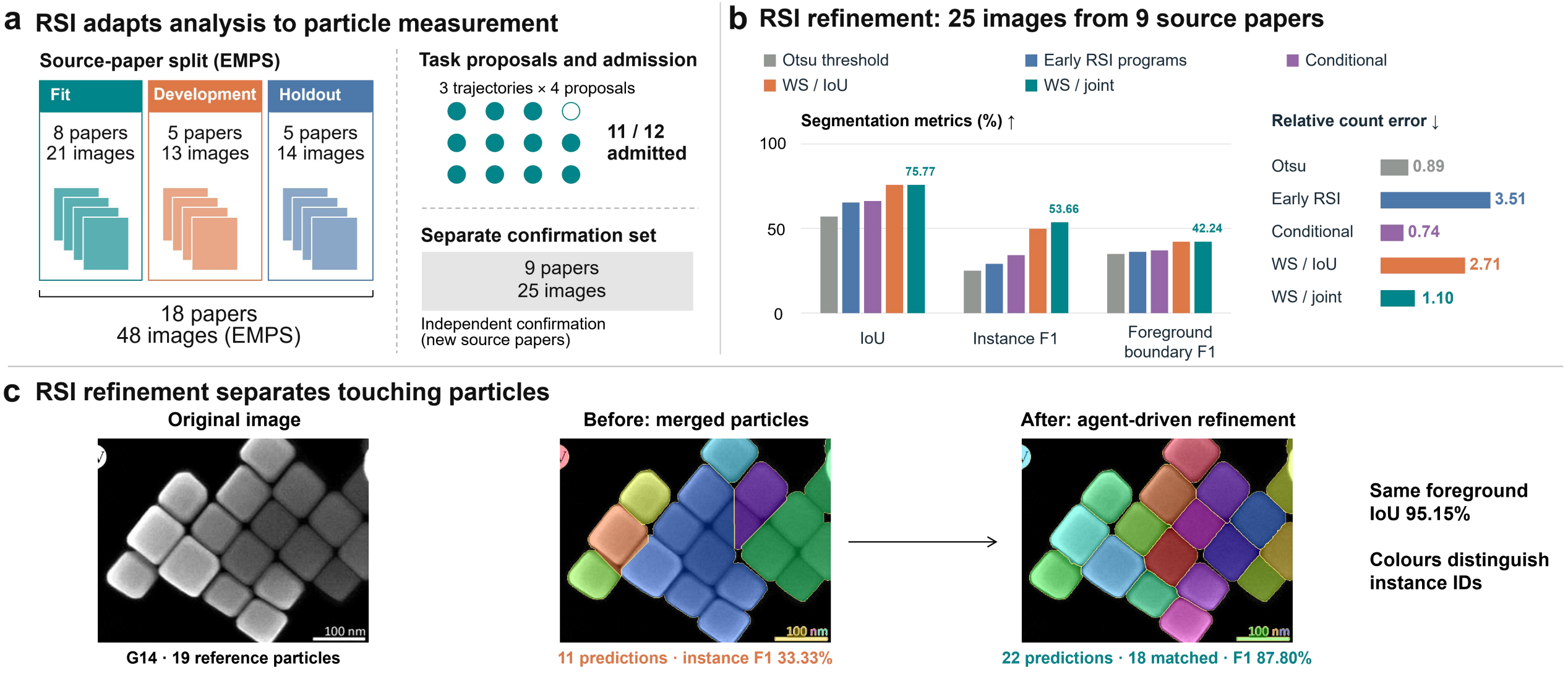

**Fig. 5 | Cross-task adaptation to nanoparticle instance separation.** a, EMPS source papers are assigned to fitting, development and holdout sets; 11 of 12 proposals across 3 trajectories are admitted. b, Confirmation of early RSI programs and predefined confi gurations on 25 images from 9 additional source papers; joint-endpoint watershed selection (WS / joint) achieves an instance F1 of 53.66%. c, Feedback-triggered, agent-driven refinement on the diagnostic G14 image separates touching particles, increasing insta nce F1 from 33.33% to 87.80% while preserving foreground IoU at 95.15% (colours distinguish instance IDs; scale bars, 100 nm).

# Discussion

PhyEvo-Mic-RSI addresses a central challenge at the intersection of computational microscopy and scienti fic agents: turning domain experts' one-off judgements and corrections into computational capabilities that an an alysis system can reuse across research iterations. Rather than program editing alone, the methodological contri bution lies in using physical evidence and expert-defined measurement targets to diagnose failure modes, evalua te proposed changes under fixed budgets, and selectively inherit useful changes across executable operators ($B_t$), learned states ($W_t$), adoption rules ($A_t$) and research lineage ($E_t$). Existing microscopy deep learning models an d tool-calling agents often rely on fixed operators and one-way task plans, requiring renewed manual interventio n when confronted with cross-material contrast reversal or atypical interfaces. PhyEvo-Mic-RSI instead synthesi zes, evaluates and inherits analysis programs and routing rules in response to specific failure modes, while equal -quota ablations identify redundant implementation details. This enables materials researchers to delegate operat or tuning, branch composition and cross-source retention checks to a closed-loop system, lowering the barrier to quantitative analysis of complex images and improving the efficiency and reproducibility of scientific character ization.

One potential limitation is that the initial physical operators, crystallographic prior templates and task-admi ssion protocols still require domain knowledge to construct and specify. Although training-free physical branche s and fixed-budget admission provide structural references and guard against capability loss, manually defining i nitial feature fields and evaluation metrics can constrain expansion to new imaging modalities, such as in situ ti me-series microscopy, four-dimensional scanning transmission electron microscopy (4D-STEM) or multimodal spectroscopic imaging. Automated extraction of operator metadata from microscopy literature or open-source al gorithm libraries, and synthesis of task-specific evaluation functions from expert-annotated examples, could ma ke capability expansion more flexible. To lower the engineering burden of extension and cross-team reuse, the f ramework provides a standardized inference interface with nine aligned fields, including the original image, phy sical responses, student predictions, variation fields and structural-domain labels. Third-party physical operators and visual models can therefore be integrated with minimal changes. An accompanying interactive review webs ite supports direct geometric correction, uncertainty marking and versioned saving, allowing experimental materials scientists without programming experience to contribute domain judgement to the self-improvement cycle.

By separating open-ended program search in the research layer from lightweight local execution in the inference layer, PhyEvo-Mic-RSI combines methodological flexibility with practical deployment. Importantly, all three strategies retained access to historical information; thus, this comparison evaluates constraints on program generation rather than inheritance itself. The results suggest that effective state inheritance need not impose rigid constraints on candidate program structures. Meanwhile, exclusion-zone supervision distils physical priors and sparse expert corrections into a compact student network with less than 2% of the parameters of retrained U-Net or EMSeek baselines, supporting fast, deterministic local inference without language-model calls during routine analysis. Crucially, the nanoparticle study demonstrates that extending the framework to a distinct microscopy task requires more than optimizing image overlap: aligning evaluation with particle identities, boundaries and counting accuracy enables research feedback to guide more meaningful analytical improvements.

Looking ahead, this stateful evolution paradigm opens several avenues for autonomous characterization and materials discovery. First, because routine inference executes locally and deterministically while program synthesis runs asynchronously in isolation, the active state $\{B_t, W_t, A_t\}$ can be coupled directly with microscope hardware controllers to guide real-time region-of-interest selection and adaptive acquisition during high-throughput or in situ experiments. Second, encapsulating analytical know-how as versioned, verifiable states allows specialized operators and routing rules evolved in one laboratory to be shared, audited and inherited across instruments and research groups without transferring raw proprietary micrographs. Finally, integrating domain-tuned open-source reasoning models with broader multimodal physical-operator libraries—and linking evolved microscopy outputs, including defect locations, interface geometries and particle-instance statistics, to downstream materials modelling and experimental validation—will move computational microscopy beyond static post-processing scripts towards self-evolving discovery platforms that accumulate human scientific expertise.

# Methods

### 1. Study design and data sources

The experiments comprised expert-supervised student learning, point-adoption optimization, program-level interface refinement and cross-task nanoparticle instance separation. The expert queue contained 154 cases; 100 corrected point cases and 51 corrected interface cases were used for the student comparisons. Interface refinement used $ZrO_2$, Al orientation I and Al orientation II, comprising 1, 30 and 20 image planes, respectively[22, 44]. $WSe_2$ images and author-provided point references came from EMSeek. Particle images and native instance masks came from EMPS. Dataset roles and fitting permissions were defined separately for each experiment; results from different protocols were not pooled into a single accuracy.

The capability-evolution stages summarized in Fig. 2c and Fig. 3a (top) represent successive RSI iterations that progressively evolved student-training configurations, hybrid fusion and geometric routing into the shared 18-branch intermediate state. In the earlier stages (Fig. 2c), the self-improvement loop explored source-specific student-learning settings (including a batch size of 32 and comparisons of 400, 800 and 1,200 training steps) and routing rules; these stages therefore trace cumulative capability evolution rather than a fixed-hyperparameter architecture comparison. The final evolution stage, selected for detailed multi-trajectory lineage and ablation analysis, began from the shared 18-branch checkpoint. During this final stage, student-network outputs were held frozen so that candidate program updates and optional numerical fitting modified ranking or combination states without retraining the deep detector at each round.

## 2. Physical evidence and executable system state

The inherited state $S_t = \{B_t, W_t, A_t, E_t\}$ comprises executable analysis capabilities, learned parameters, observable adoption rules and experimental experience, respectively. Inference reads the active capabilities, weights and adoption rules without expert labels. Research evaluates proposed changes before updating the active parent; rejected proposals contribute records but do not replace the parent. Subsequent proposals receive the active parent and permitted earlier outcomes.

For interface analysis, local patches were mean-centred, variance-normalized and Hann-windowed. Fourier power was binned radially and by orientation; square-root power descriptors were normalized to unit length. Channels were standardized by their median and 1.4826 times the median absolute deviation, clipped to [−1.5, 1.5], and partitioned with iterative median prototypes. Upsampled assignments formed a structural-domain map; the sum of descriptor-gradient norms along the two image axes formed the variation field. These are structural-texture measurements, not chemical-phase assignments.

The frozen 18-branch system selected or combined spatial rankings from observable support fractions, variation distributions and branch responses. Candidate programs received nine aligned fields: the image, base score, candidate support, general validity, physical response, student response, physical validity, variation and structural-domain labels. Programs returned a finite score array and Boolean support of the original image shape; selection outside the general valid region was prohibited. Expert geometries and scoring-source identities were unavailable to inference.

## 3. Expert supervision and student training

Two experts divided the queue between them, inspecting original images alongside teacher predictions and accepting, correcting or rejecting the proposed geometry. This was prediction-assisted review, not independent duplicate annotation. The final records contained 152 corrections (100 point cases, 51 interface cases and 1 disordered-region case), 1 acceptance and 1 rejection. The point cases contained 808 explicit positives. Positive annotations from the 21 cases with empty teacher predictions were retained independently of those predictions.

Human points were rasterized as positive discs of radius 3 pixels; closed interface geometries became positive bands. Unlabelled regions remained unknown and contributed no human-negative loss. Teacher supervision was masked within four binary dilations of human positives, preventing conflicting weak teacher-background signals from suppressing new targets. Empty or abstaining teacher predictions did not generate image-wide background labels. Each supervision term was normalized within each image and then averaged over images containing that term.

The student was a U-Net-like encoder–decoder with base channel width 16 and 119,378 parameters; evidence and reliability heads without corresponding objectives were frozen. Images were scaled using the 0.5th and 99.5th intensity percentiles and clipped to [0, 1]. The fixed comparison used random initialization, 128 × 128-pixel patches, batch size 16, Adam with learning rate 0.001, gradient-norm clipping at 5 and 800 steps, with seeds 23, 41 and 67. Methods shared the permitted 151 cases and crop/augmentation schedules. No $WSe_2$ images were used to fit student weights.

Let $L_{\mathrm{H}}$ be binary cross-entropy on human positives, $L_{\mathrm{T}+}$ and $L_{\mathrm{T}-}$ the teacher-positive and weak-background terms, $q$ the mean predicted probability over valid pixels, and $c$ the task occupancy ceiling. The joint-supervision loss was

$$L = 2L_{\mathrm{H}} + 2\max(q - c, 0)^2 + L_{\mathrm{T}+} + 0.25L_{\mathrm{T}-}$$

with $c = 0.03$ for point tasks and $c = 0.15$ for interfaces. The expert-only control omitted the two teacher terms but retained the occupancy regularizer, which did not label unknown pixels as human negatives. U-Net (7,765,541 parameters) and EMSeek Ref-UNet (27,677,461 parameters) were trained from scratch under the same step budget. Ref-UNet averaged responses from at most three positive training-reference patches; no test-image references were used. Equal steps did not imply equal computation.

## 4. $WSe_2$ adoption rules and point evaluation

The 107 unique $WSe_2$ files were ordered deterministically by hashing a fixed prefix with each image SHA-256, then assigned to fitting (32), proposal selection (22) and final confirmation (53). They were treated as one known collection, not independent acquisition sources. Frozen physical candidates and three-seed student predictions were reused. Only fitting labels trained the selector; selection feedback determined admission, and confirmation was performed after the program and numerical state were frozen.

A regularized logistic selector combined physical ranks and scores, distances to student peaks, student responses and local contrast, image intensity, multiscale residuals, Laplacian responses and branch agreement. Features were robustly standardized, loss was balanced by image, and calibration used four-fold image-level out-of-fold predictions within the fitting set. Final coordinates came only from the physical-candidate pool; student evidence altered adoption rather than adding student-only points.

The top 60% of baseline candidates were protected. At most 10% were replaced from a pool capped at 1.25 times the original budget. A replacement required probability ≥0.55, an advantage ≥0.15 over the displaced point, and the minimum inter-point distance. Tail removal used a 0.32 threshold; its cap rose from 10% to 15% only after an incremental fitting-set gate passed. Admission required a three-seed mean selection-F1 increase of at least 0.002, with no seed declining by more than 0.005 relative to the parent or original physics. Fixed fusion used 50:50 candidate quotas. Same-label physics-only calibration used the same program and labels but replaced variable student evidence with constants during fitting and inference.

For an $H \times W$ image, the point cap was $K = \max\left(1, \text{round}\left(\frac{50HW}{512^2}\right)\right)$, with student peaks at least 4 pixels apart. Reference points were centroids of four-connected non-black components in author annotation images. One-to-one bipartite matching used a 6-pixel Euclidean tolerance, maximizing the number of matches before minimizing distance. Matched predictions defined TP, unmatched predictions FP and unmatched references FN:

$$P = \frac{\text{TP}}{\text{TP} + \text{FP}}, \quad R = \frac{\text{TP}}{\text{TP} + \text{FN}}, \quad F_1 = \frac{2\text{TP}}{2\text{TP} + \text{FP} + \text{FN}}$$

Counts were pooled across images before scoring each seed, then metrics were averaged across seeds. The 107-image transfer experiment and 53-image rule confirmation were scored separately. The latter contained 7,969 reference points. Sparse expert annotations in the five-source development study supported known-positive coverage, not exhaustive precision over unlabelled regions.

## 5. Interface coverage and RSI admission

For plane $i$, let $G_i$ denote expert-positive pixels. Pixels with positive scores within valid candidate support were ranked in stable descending order, selecting at most $K_i = \lceil 0.10 H_i W_i \rceil$ pixels to form $Q_i$. Plane coverage, source coverage and the three-source objective were

$$C_i = \frac{|Q_i \cap G_i|}{|G_i|}, \quad C_s = \frac{1}{|I_s|} \sum_{i \in I_s} C_i$$

$$J = \frac{C_{\mathrm{ZrO_2}} + C_{\mathrm{Al\,I}} + C_{\mathrm{Al\,II}}}{3}$$

where $I_s$ is the set of planes belonging to source $s$. Thus $J$ is equal-source positive coverage at fixed area, not pixel accuracy or IoU. Both the capability-development stages and subsequent RSI trajectories were evaluated using this plane-averaged, equal-source objective $J$.

Proposals specified observations, a mechanistic hypothesis, executable changes, applicability conditions and a falsifiable expectation, with optional numerical fitting. New fitting was allowed only on the 20 Al orientation II planes. $ZrO_2$ and Al orientation I were excluded from fitting, but their evaluation feedback was visible during development. Independent proposal sessions received anonymized diagnostics, the active parent, permitted historical programs and measured feedback—not raw images, complete annotations or subsequent historical solutions. Generated programs ran in network-disabled isolated processes; only the fitting interface exposed permitted labels.

Admission required $\Delta J \geq 0.002$ relative to the current parent and the same minimum mean gain across the two fit-excluded sources. These fractional gains correspond to 0.2 percentage points. No source could regress by more than 0.002 against the frozen, current or previously retained references. A 90° rotation test imposed the same tolerance relative to untransformed performance and retained rotated references; reverse-order execution of all 51 planes then had to reproduce scores, support and selections. Invalid outputs, timeouts and execution failures were rejected. Admitted code and fitted state became the next actual parent; rejection retained the existing parent.

The main searches had 10, 12 and 14 proposal opportunities; three single-step routing references were recorded separately. Requests specified GPT-5.6 Sol with xhigh reasoning, while the server-side backend version was not independently verified. Calls were session-orchestrated within predefined execution and evaluation protocols; researchers maintained the initial objectives and interfaces.

## 6. Inheritance checks and mechanistic interventions

Inheritance was checked through actual ancestor-program calls, numerical-state reads and replayed outputs. Three retained terminal programs were replayed on all 51 original field packages (153 comparisons). Horizontal reflection and 180° rotation were additionally tested for three program seeds, three branch types (frozen, routing and full) and 51 planes, giving 918 predictions. These checks tested execution reproducibility and transformation consistency, not additional independent acquisitions.

For expert positives $G$, the gained and lost positives between parent and updated selections were

$$N_{\mathrm{gain}} = \left|\left(Q_{\mathrm{new}} \setminus Q_{\mathrm{parent}}\right) \cap G\right|$$

$$N_{\mathrm{loss}} = \left|\left(Q_{\mathrm{parent}} \setminus Q_{\mathrm{new}}\right) \cap G\right|$$

and the net gain was $N_{\mathrm{gain}} - N_{\mathrm{loss}}$. Equal-exchange interventions fixed both total selected area and the number of pixels introduced from outside the parent selection. The three boundary-shell transitions used an exchange quota of 1,259 pixels; the remaining quota was filled using parent scores. Domain correction, inverse-residual evidence, inherited-parent state, component gating and cross-scale aggregation were intervened on individually without refitting coefficients or reselecting configurations.

All 8 admitted nodes were tested on 51 planes. The proximity control matched support, selected-pixel count and exchange quota, ranking added pixels by distance to the parent selection and breaking ties by parent score and stable index. It tested the added ranking mechanism within the same support, not how that support was discovered. Gains, ties and declines were all retained. Parent-reset controls changed only the current root's direct parent call to the frozen system, preserving other inputs and coefficients.

**7. Matched-opportunity search strategies**

Structured inheritance, generic iterative coding and structured replacement were compared using three trajectories per strategy and twelve opportunities per trajectory (108 total). All groups shared the 18-branch starting system, input fields, Al II fitting permissions, anonymized feedback and admission rules. Structured inheritance required complete research fields and callable parents. Generic coding retained the same evidence and history interfaces but relaxed proposal structure. Replacement retained access to historical code and numerical state without automatic parent-call dispatch, permitting earlier computations to be copied or inlined.

Evaluation followed a fixed interleaved schedule with identifiers 23, 41 and 67. Each fitting or inference stage was limited to 900 seconds; fitting was capped at 4,096 steps and numerical state at 512 KiB before and after compression. A failure consumed one opportunity, without coordinator repair or a free retry, and left the parent score unchanged. All accepted, rejected and failed proposals contributed to the totals. Equal opportunities did not imply equal actual token usage, computation or cost.

Research-strategy evolution was evaluated separately from the 108-opportunity comparison. Original and revised strategies were compared from two matched starting states under fixed proposal opportunities and evaluation criteria. Strategy admission was based on the predefined mean non-negative retained gain per proposal opportunity. Admitted strategies were subsequently used to guide program generation, while analysis-program updates remained subject to the original admission criteria.

**8. Particle adaptation, endpoint selection and boundary refinement**

EMPS images were grouped by source-paper DOI before scoring, excluding sources used in preliminary examples and checking exact duplicates and feature overlap. The initial split contained 21 fitting images from 8 papers, 13 development images from 5 papers and 14 holdout images from 5 papers. A further 25 images from 9 papers were selected in the same registered source order for confirmation. Entire papers remained within one split; paper grouping did not establish original acquisition identity. Native instance masks were the references.

Three four-proposal trajectories started from Otsu thresholding. Candidates received raw grayscale images and returned foreground or instance masks; Boolean masks were converted using eight-connectivity. Initial admission required a development-source-averaged foreground-IoU gain >0.0001. Later confirmation compared twelve archived programs with twelve-configuration conditional and marker-controlled watershed libraries. Conditional selection combined global/local contrast, occupancy and minimum-area rules; watershed used thresholded foreground, polarity assessment, distance-derived markers and at least one marker per component. Library order and selection rules were fixed before confirmation.

Foreground IoU measured intersection over union. Foreground boundary F1 used a 2-native-pixel tolerance; instance F1 used one-to-one matching at instance IoU ≥0.5. Absolute count error and relative count error were

$$e_{\mathrm{abs}} = \left|N_{\mathrm{pred}} - N_{\mathrm{ref}}\right|, \quad e_{\mathrm{rel}} = \frac{\left|N_{\mathrm{pred}} - N_{\mathrm{ref}}\right|}{\max\left(N_{\mathrm{ref}}, 1\right)}$$

where $N_{\mathrm{pred}}$ and $N_{\mathrm{ref}}$ are predicted and reference counts. Metrics were averaged first within each source paper and then equally across papers. IoU-only selection required a gain >0.0001; joint admission required non-decreasing IoU, instance-F1 gain >0.0001 and non-increasing relative count error, using numerical tolerance $10^{-12}$. Development-IoU ties were retained in interpreting comparisons rather than treated as proof of universal superiority over every tie-breaking rule.

For the previously seen G14 diagnostic image, a feedback-triggered, agent-driven refinement was conducted outside the three fixed four-proposal trajectories. Following qualitative user feedback that the predicted boundaries were implausible, the research agent autonomously diagnosed the failure of distance-only partitioning and implemented image-derived particle-centre markers with intensity-guided watershed topography (evaluating nine centre-detection configurations), followed by a second-stage decoupling of coarse centre-smoothing from fine seam-tracing scales (evaluating two boundary-scale configurations) while preserving the foreground mask. All configurations were selected on the original development set using instance F1 and internal-seam boundary F1—without constraining predictions by the known G14 particle count—retaining centre-scale and fine-topography coefficients of 0.012 and 0.0015 (prominence, 0.05) and verifying stability via isolated replay and regression checks. This interactive session-level refinement was reported separately from the frozen new-source confirmation.

### 9. Statistical summaries and reproducibility

Point F1 was calculated from pooled counts, interface coverage from equal-source averages, and nanoparticle segmentation metrics from equal-paper averages. Task-specific metrics were reported separately rather than combined into a single performance score. Seed means describe training or execution variability, not independent material acquisitions; pixels and adjacent image planes were not treated as independent replicates for significance inference. Input versions, programs, fitted states and decision records were linked for reproducibility, with admitted states distinguished from rejected or failed trials.

## Acknowledgments

This work was financially supported by the National Natural Science Foundation of China (No. 22322203 and 2 2275110),  the Key Research and Development Program of Inner Mongolia and Ordos (Ordoslab-kjzc-202603-01, Ordoslab-kjzc-202506, Ordoslab-sysjc-202502), the HE science foundation and Central Universities Young Faculty Research Innovation Capacity Support Program (SRICSPYF-ZY2025039), the National Natural Scienc e Foundation of China (Nos. 52650103, 52322212).

## Author contributions

Y.Y. conceived the original research idea, developed and coordinated the research plan, conducted the literature review and data collection, developed and implemented the entire framework, performed the computational exp eriments and evaluated the results, prepared the figures, and wrote the original manuscript. S.T. and Y.Z. perfor med the expert annotations of electron microscopy images. S.T. and J.Q. also contributed to the literature review and manuscript revision. F.W., Q.Z., and X.C. provided scientific guidance, participated in research discussions, and critically reviewed and revised the manuscript, offering constructive suggestions. All authors reviewed and approved the final manuscript.

## Competing interests

Authors declare that they have no competing interests.